\documentclass{article} %
\usepackage{iclr2025_delta,times}

\usepackage{amsmath}
\usepackage{mathtools}
\usepackage{booktabs}
\usepackage{array}
\usepackage{comment}
\usepackage[small]{caption}

\definecolor{darkblue}{HTML}{003095}
\usepackage[colorlinks=true, linkcolor=darkblue, urlcolor=darkblue, citecolor=darkblue, anchorcolor=darkblue]{hyperref}

\usepackage[capitalise]{cleveref}

\usepackage{amssymb}
\usepackage{wrapfig}
\usepackage{subcaption,siunitx,booktabs}
\usepackage{bm}

\newcommand{\bepsilon}{\bm{\epsilon}}
\newcommand{\btheta}{\bm{\theta}}
\newcommand{\bigeta}{\bm{\eta}}
\newcommand{\z}{\mathbf{z}}
\newcommand{\x}{\mathbf{x}}
\newcommand{\s}{\mathbf{s}}
\newcommand{\stheta}{\s_{\btheta}}
\newcommand{\stildetheta}{\tilde{\s}_{\btheta}}
\newcommand{\htheta}{\mathbf{h}_{\btheta}}
\newcommand{\w}{\mathbf{w}}
\newcommand{\wtilde}{\tilde{\w}}

\newcommand{\0}{\mathbf{0}}
\newcommand{\I}{\mathbf{I}}

\newcommand{\Normal}{\mathcal{N}}
\newcommand{\diff}{\mathrm{d}}
\newcommand{\dt}{\diff t}
\newcommand{\dz}{\diff\z}
\newcommand{\dx}{\diff\x}
\newcommand{\dw}{\diff\w}
\newcommand{\dwtilde}{\diff\wtilde}
\newcommand{\Ldiff}{\mathcal{L}_{\mathrm{diff}}}
\newcommand{\ptheta}{p_{\btheta}}
\newcommand{\R}{\mathbb{R}}
\newcommand{\E}{\mathbb{E}}
\newcommand{\KL}{D_{\mathrm{KL}}}
\newcommand{\qdata}{q_{\mathrm{data}}}

\title{Breaking the Likelihood--Quality Trade-off\\in Diffusion Models by Merging\\Pretrained Experts}

\author{
  Yasin Esfandiari\textsuperscript{1}\thanks{Work done while at Helmholtz AI. Correspondence to \href{mailto:yaes00001@stud.uni-saarland.de}{{yaes00001@stud.uni-saarland.de}}.} \quad
  Stefan Bauer\textsuperscript{2,3} \quad
  Sebastian U. Stich\textsuperscript{4} \quad
  Andrea Dittadi\textsuperscript{2,3,5} \\
  \textsuperscript{1}Saarland University \quad
  \textsuperscript{2}Helmholtz AI  \quad
  \textsuperscript{3}Technical University of Munich  \\
  \textsuperscript{4}CISPA Helmholtz Center for Information Security \quad 
  \textsuperscript{5}MPI for Intelligent Systems, Tübingen
}

\iclrfinalcopy %
\begin{document}

\maketitle

\begin{abstract}
Diffusion models for image generation often exhibit a trade-off between perceptual sample quality and data likelihood: training objectives emphasizing high-noise denoising steps yield realistic images but poor likelihoods, whereas likelihood-oriented training overweights low-noise steps and harms visual fidelity. We introduce a simple plug-and-play sampling method that combines two pretrained diffusion experts by switching between them along the denoising trajectory. Specifically, we apply an image-quality expert at high noise levels to shape global structure, then switch to a likelihood expert at low noise levels to refine pixel statistics. The approach requires no retraining or fine-tuning---only the choice of an intermediate switching step.
On CIFAR-10 and ImageNet32, the merged model consistently matches or outperforms its base components, improving or preserving both likelihood and sample quality relative to each expert alone.
These results demonstrate that expert switching across noise levels is an effective way to break the likelihood–quality trade-off in image diffusion models.
\end{abstract}

\section{Introduction}

Diffusion models are a class of probabilistic generative models that learn to approximate a data distribution by reversing a forward noising process through a learned denoising procedure~\citep{sohl2015deep, ho2020denoising, nichol2021improved}. They have recently achieved state-of-the-art results, e.g., in image generation~\citep{dhariwal2021diffusion, tang2024generative, kim2024pagoda}, density estimation~\citep{kingma2021variational}, and in text-to-image and text-to-video generation tasks~\citep{esser2024scaling, polyak2024movie}.

For image data, likelihood and perceptual quality are often misaligned in practice~\citep{theis2015note}, that is, strong performance on one does not necessarily imply good performance on the other. Notably, \citet{kim2021soft} report an inverse correlation between likelihood and sample quality as measured via the Fréchet Inception Distance (FID). As a result, models that aim to maximize likelihood typically optimize a lower bound on it, whereas models prioritizing perceptual quality modify the training objective, e.g., by reweighting contributions from different time steps in the diffusion process. This trade-off implies that models producing visually appealing samples often achieve lower likelihoods, while those optimized for likelihood tend to generate less realistic images. Because likelihood and FID capture complementary aspects of generative modeling---the statistical fidelity of the data versus its perceptual realism---balancing both is crucial for developing diffusion models that accurately represent the data distribution while producing convincing visual samples.

In this paper, we aim to overcome the likelihood--FID trade-off by designing a model that can generate images with both high perceptual quality and strong likelihood. To do this, we start from two key empirical observations reported in the literature:
(1) Higher noise levels are associated with perceptual image quality. For example, DDPM~\citep{ho2020denoising} employs a simplified objective that down-weights the loss at lower noise levels, allowing the model to focus on the more challenging denoising steps at higher noise levels. Similarly, \citet{kim2021soft} showed that accurate score prediction at high noise levels is crucial for generating realistic samples. 
(2) Likelihood is highly sensitive to low-level pixel statistics~\citep{zheng2023improved, kim2021soft}, whereas perceptual quality is primarily determined by global image structure rather than fine-grained pixel details. Supporting this view, \citet{kingma2018glow} and \citet{kingma2024understanding} showed that training on 5-bit images, which effectively discards fine details, can lead to better visual quality.

\begin{figure}
\centering
\includegraphics[width=0.65\textwidth]{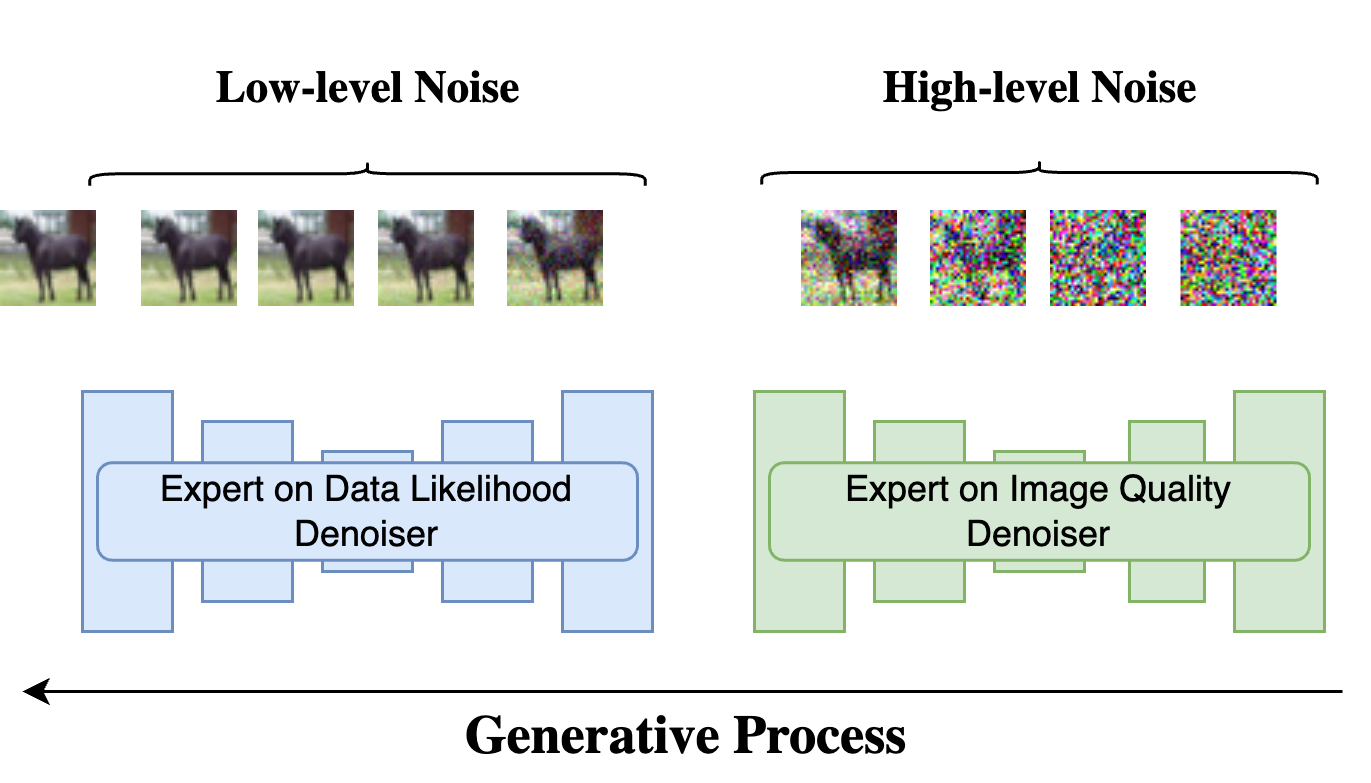}
\caption{Diagram of our merged model where at an intermediate time $\eta \in [0,1]$ we switch between denoisers. Note that the likelihood model is only used for almost imperceptible noise levels. This significantly improves the likelihood, which is sensitive to low-level color statistics, while leaving the FID unaffected.}
\label{fig:merged_model}
\end{figure}

Motivated by these insights, we propose a simple approach that merges two pretrained diffusion experts---one specialized in image quality and the other in likelihood. Specifically, we use EDM~\citep{karras2022elucidating} as the image-quality expert for high noise levels, and VDM~\citep{kingma2021variational} as the likelihood expert for low noise levels. An overview of the merged model is provided in \cref{fig:merged_model}. 
Starting from noise, the model first denoises up to a chosen intermediate step using the image-quality expert, producing a high-fidelity yet slightly noisy sample. The process then switches to the likelihood expert, which refines the sample to improve likelihood while preserving perceptual quality. By appropriately selecting the switching point along the denoising trajectory, the model achieves strong performance in both FID and likelihood, effectively overcoming the trade-off between the two.

The remainder of this paper is organized as follows. \Cref{sec:background} introduces the necessary preliminaries on diffusion models. \Cref{sec:methodology} presents our framework for adapting pretrained models for reuse across different processes, enabling the merging of multiple experts. \Cref{sec:experiments_results} describes the experimental setup and reports quantitative and qualitative results on CIFAR-10 and ImageNet32. Finally, \Cref{sec:related} discusses related work, and \Cref{sec:conclusion} concludes with limitations and directions for future research.

\section{Preliminaries}
\label{sec:background}

Diffusion models~\citep{sohl2015deep, song2019generative, ho2020denoising} are a class of generative models that learn to reverse a diffusion process that gradually perturbs data with noise.
Let $\x \in \R^d$ denote a data point drawn from an unknown distribution $\qdata$. The forward process is a continuous-time stochastic process $(\z_t)_{t\in[0,1]}$ in $\R^d$ initialized from a simple conditional distribution $q(\z_0 | \x) = \mathcal{N}(\z_0; \alpha_0 \x, \sigma_0^2 \I)$ with scalar parameters $\alpha_0,\sigma_0$.
A common choice for the forward dynamics is an Ornstein--Uhlenbeck SDE with a time-dependent but data-independent drift:
\begin{align}
    \dz_t = f_t \z_t \dt + g_t \dw_t \ ,
\end{align}
where $\w_t$ is a standard Wiener process and $f_t,g_t$ are scalar functions of time.
Under this construction, the marginal density of $\z_t$ conditional on data is
\begin{align}
    q(\z_t | \x) = \mathcal{N}(\z_t; \alpha_t \x, \sigma_t^2 \I) \ ,
\end{align}
where $\alpha_t, \sigma_t \in \mathbb{R}_{>0}$ are smooth scalar-valued functions of $t$ defining the \emph{noise schedule}. %
Choosing $(\alpha_t,\sigma_t)$ determines the SDE coefficients via:
\begin{equation}
    f_t = \frac{\mathrm{d} \log \alpha_t}{\mathrm{~d} t} \ , \qquad 
    g_t^2 = 
    \alpha_t^2 \; \frac{\diff}{\dt}\!\left[\frac{\sigma_t^2}{\alpha_t^2}\right] \ .
    \label{eq:def_vdm2sde}
\end{equation}
We assume that the \textit{signal-to-noise ratio} (SNR), $\alpha_t^2/\sigma_t^2$, is monotonically decreasing in $t$, which makes the diffusion coefficient $g_t$ from \cref{eq:def_vdm2sde} well defined.

If we could sample from $q(\z_1)$ and the marginal scores $\nabla_{\z_t} \log q(\z_t)$ were known, we could obtain samples from $q(\z_0)$ by simulating a deterministic or stochastic process backward in time. In the stochastic settings, several reverse-time SDEs are possible, with different diffusion coefficients $g_t$. A common choice uses the same $g_t$ as in the noising process~\citep{song2020score}:
\begin{equation}
\dz_t = [f_t \z_t - g_t^2 \nabla_{\z_t} \log q(\z_t)] \dt + g_t \dwtilde_t \ , \qquad \z_1 \sim q(\z_1) \ ,
\label{eq:background:reverse_diffusion}
\end{equation}
where $\wtilde_t$ denotes a reverse-time Wiener process. 
Another common alternative is to define a deterministic process known as the \emph{probability flow ODE} (PF~ODE)~\citep{song2020score}:
\begin{equation}
  \frac{\dz_t}{\dt} = f_t \z_t -\frac{1}{2} g_t^2 \nabla_{\z_t} \log q(\z_t)  \ , \qquad \z_1 \sim q(\z_1) \ .
  \label{eq:background:pf_ode}
\end{equation}
Both \cref{eq:background:reverse_diffusion,eq:background:pf_ode} share the same time-marginals $q(\z_t)$ as the forward process and therefore yield exact samples from $q(\z_0)$. Generating data then requires a decoder $q(\x|\z_0) \propto q(\z_0|\x)\qdata(\x)$ which is typically unavailable.

In practice, we replace the intractable ingredients by:
(i) a \emph{prior} $p(\z_1) \approx \int q(\z_1|\x)\qdata(\x)\mathrm{d}\x$, usually $\Normal(\0, \I)$;
(ii) a \emph{score estimator} $\stheta(\z_t, t) \approx \nabla \log q(\z_t)$;
(iii) a \emph{likelihood function} $p(\x|\z_0) \approx q(\x|\z_0)$, e.g., chosen to be proportional to $q(\z_0|\x)$~\citep{kingma2021variational}.
By using these three approximations to define the generative model, sampling a new data point $\x$ proceeds by (i) drawing $\z_1 \sim p(\z_1)$, (ii) integrating \cref{eq:background:reverse_diffusion} or \cref{eq:background:pf_ode} with the approximate score $\stheta$ to obtain $\z_0$, and (iii) sampling $\x \sim p(\x|\z_0)$.

The generative model can be learned by minimizing an upper bound on the negative log-likelihood:
\begin{align}
    \label{eq:def_vlb_likelihood}
    -\log \ptheta(\x) &\leq \underbrace{\KL(q(\z_1 | \x) \| p(\z_1))}_{\text{Prior loss}} + \underbrace{\E_{q(\z_0 | \x)}[-\log p(\x | \z_0)]}_{\text{Reconstruction loss}} + \underbrace{\Ldiff(\x; \btheta)}_{\text{Diffusion loss}} \\  %
    \Ldiff(\x; \btheta) &\coloneqq \frac{1}{2} \E_{t \sim \mathcal{U}(0,1), \z_t \sim q(\z_t|\x)} \left[g_t^2 \, \left\| \stheta(\z_t, t) - \nabla \log q(\z_t|\x)\right\|^2\right] \ .
\end{align}
In this setting, the noising process, the prior distribution, and the likelihood $p(\x|\z_0)$ are fixed. Learning the generative model then amounts to learning the score estimator by minimizing $\Ldiff$ over the training data. The minimizer of this objective corresponds to the \emph{marginal} score $\nabla \log q(\z_t)$ \citep{vincent2011connection}. This result provides a theoretical justification for the score-based approach: although the model is trained via a tractable regression objective against the conditional score, the procedure yields an accurate estimate of the desired marginal score. Consequently, the training objective not only aligns with maximum likelihood principles but also remains computationally efficient and theoretically well grounded.
Moreover, this objective can be equivalently reformulated through various parameterizations of the score, such as predicting the original data, the added noise, the velocity \citep{salimans2022progressive}, or the PF~ODE vector field \citep{lipman2022flow,liu2022flow}---see, e.g., \citet{kingma2024understanding} for an overview.

Besides enabling deterministic sampling from diffusion models, the PF~ODE \eqref{eq:background:pf_ode} allows us to compute a tighter bound than \cref{eq:def_vlb_likelihood}. The exact likelihood on any $\z_0$ can be computed via the instantaneous change of variables formula \citep{chen2018neural}: 
\begin{align}
    \log \ptheta(\z_0) & =\log p(\z_1) + \int_0^1 \nabla \cdot \htheta (\z_t, t) \dt 
    \label{eq:def_ode_likelihood}
\end{align}
where $\htheta (\z_t, t)$ is the vector field of the PF~ODE \eqref{eq:background:pf_ode} with $\nabla \log q(\z_t)$ replaced by $\stheta(\z_t, t)$.
The data log-likelihood can then be bounded as follows:
\begin{align}
    \log \ptheta(\x) \geq \E_{q(\z_0|\x)} \left[ \log p(\x|\z_0) + \log \ptheta(\z_0) - \log q(\z_0 | \x) \right] \ ,
    \label{eq: ODE ELBO}
\end{align}
which can be estimated by Monte Carlo sampling since $\log \ptheta(\z_0)$ can be computed with \cref{eq:def_ode_likelihood} and the other two terms can be designed to be tractable. The bound gap is exactly $\KL(q(\z_0|\x) \| p(\z_0|\x))$, which is typically negligible in practice.
When $q(\z_0|\x)$ is a \emph{dequantization} distribution \citep{ho2019flow,zheng2023improved}, $p(\x|\z_0) = 1$ almost surely under $q(\z_0|\x)$, and thus the $\log p(\x|\z_0)$ term in \cref{eq: ODE ELBO} vanishes.

\section{Merging Experts}
\label{sec:methodology}

In this section, we discuss the problem of \emph{merging} multiple pretrained diffusion models and demonstrate how this can be applied to our specific objective: combining two pretrained experts, each specialized in one of two fundamental aspects of generative modeling---density estimation and perceptual image quality.
Our approach is based on reformulating the diffusion dynamics in terms of the signal-to-noise ratio (SNR), which provides a natural framework for aligning and integrating models trained under different noise schedules. When considering standard diffusion or flow models defined by Gaussian probability paths, this formulation can be applied directly.
We begin by presenting a general method for adapting a pretrained score model to a new stochastic process. We then show how this procedure can be used to merge the two expert models of interest---one optimized for likelihood and the other for perceptual fidelity.

Let the \emph{target forward process} be defined by data-conditional marginals $q(\z_t|\x) = \Normal(\alpha_t \x, \sigma_t^2 \I)$ with a smooth noise schedule $t \mapsto (\alpha_t, \sigma_t)$. We define its negative log-SNR as $\gamma_t \coloneqq - \log \frac{\alpha_t^2}{\sigma_t^2}$ which increases monotonically with $t$. Given a score estimate $\stheta(\z_t, t) \approx \nabla \log q(\z_t)$, new data samples can be approximately generated using the reverse SDE or the PF~ODE associated with this target process (\cref{eq:background:reverse_diffusion,eq:background:pf_ode}). In our merged model, rather than training a new score network, we employ score estimates obtained from pretrained expert models.

Each pretrained model has been trained under its own noising process with marginals $\tilde{q}(\tilde{\z}_u|\x) = \Normal(\tilde{\alpha}_u \x, \tilde{\sigma}_u^2 \I)$ and corresponding noise schedule $u \mapsto (\tilde{\alpha}_u, \tilde{\sigma}_u)$ and negative log-SNR $\tilde{\gamma}_u$. During training, the model learned a score estimator $\stildetheta(\tilde{\z}_u, u) \approx \nabla_{\tilde{\z}_u} \log \tilde{q}(\tilde{\z}_u)$ or an equivalent quantity (e.g., noise, data, velocity \citep{salimans2022progressive}, or PF~ODE vector field \citep{lipman2022flow,liu2022flow}) that can be converted to a score. The score model was trained on inputs distributed as $\tilde{q}(\tilde{\z}_u) = \int \qdata(\x) \Normal(\tilde{\z}_u \,;\, \tilde{\alpha}_u \x, \tilde{\sigma}_u^2 \I) \dx$. Our goal is to reuse this pretrained score function within the target process at any desired time $t$.

At time $t$ of the target process, the state $\z_t$ has a noise level $\gamma_t$. To use a pretrained model at this point, we match the noise levels of the two processes by equating their negative log-SNRs:
\begin{align}
    \tilde{\gamma}_u = \gamma_t \ \qquad u = \tilde{\gamma}^{-1}(\gamma_t)
\end{align}
i.e., we identify the expert's time $u$ that corresponds to the same effective noise level as that of the target process at time $t$.
This mapping is well-defined only where such a $u$ exists, i.e., when the desired value of $\gamma_t$ lies within the range $\tilde{\gamma}([0,1])$ of the expert's noise schedule.
Finally, we rescale $\z_t$ so that it follows the same distribution as the expert's training data:
\[
    \tilde{\z}_u 
    \coloneqq \frac{\tilde{\alpha}_u}{\alpha_t} \z_t
    = \frac{\tilde{\alpha}_u}{\alpha_t} (\alpha_t \x + \sigma_t \bepsilon)
    = \tilde{\alpha}_u \x + \tilde{\sigma}_u \bepsilon \ , \qquad \text{with } \bepsilon \sim \Normal(\0, \I) \ ,
\]
which is distributed as $\tilde{q}(\tilde{\z}_u)$, as required.
Now, $\stildetheta(\tilde{\z}_u, u)$ approximates $\nabla \log \tilde{q}(\tilde{\z}_u)$, whereas our goal is to approximate $\nabla \log q(\z_t)$. By a change of variables, we have:
\begin{align}
    \nabla_{\z_t} \log q(\z_t) &= \nabla_{\z_t} \log \tilde{q}\left( \tilde{\z}_u\right)  
    = \frac{\tilde{\alpha}_u}{\alpha_t} \nabla_{\tilde{\z}_u} \log \tilde{q}\left( \tilde{\z}_u\right) \ .
\end{align}
Hence, the score of the target process can be approximated by:
\begin{align}
    \stheta(\z_t, t) = \frac{\tilde{\alpha}_u}{\alpha_t} \ \stildetheta\big(\tfrac{\tilde{\alpha}_u}{\alpha_t} \z_t, \ \tilde{\gamma}^{-1}(\gamma_t) \big) \ .
\end{align}
In the experiments presented in this work, we restrict our attention to variance-preserving (VP) processes, for which $\alpha_t^2 + \sigma_t^2 = 1$ for all $t$. Under this condition, the equality of SNRs implies that $\tilde{\sigma}_u = \sigma_t$ and $\tilde{\alpha}_u = \alpha_t$. Consequently, the scaling factor cancels out and the adaptation simplifies to a straightforward time remapping:
\begin{align}
    \stheta(\z_t, t) = \stildetheta\left(\z_t, \ \tilde{\gamma}^{-1}(\gamma_t) \right) \ .
\end{align}

Having established how to adapt a pretrained expert to different noise processes, we now turn to the case involving multiple pretrained experts. By applying the adaptation procedure described above, each expert can generate a score estimate that remains valid across any noise level it was trained on. To construct a unified model, these estimates can then be combined in an ensemble fashion, e.g., by taking a convex combination of the scores from the individual models.

In this work, we focus on the special case in which a hard switch is applied between two models, i.e., we assign a weight of 1 to exactly one expert at any given time, effectively employing only a single pretrained expert at each step.
The concrete instantiation of this procedure with pretrained VDM \citep{kingma2021variational} and EDM \citep{karras2022elucidating} models is presented in \cref{sec:experiments_results}.

\section{Experiments}
\label{sec:experiments_results}

To assess the effectiveness of this hybrid approach, we conduct a series of experiments combining the two pretrained diffusion experts through a hard switch during denoising. We first describe the experimental configuration, then analyze quantitative and qualitative results demonstrating how the switching threshold mediates the trade-off between perceptual quality and data likelihood.

\subsection{Experimental Setup}
\label{sec: experimental setup}

We use pretrained EDM~\citep{karras2022elucidating} and VDM~\citep{kingma2021variational} models and define a hard switch between them during the denoising process. The EDM model operates at high noise levels $\gamma_t$ to enhance perceptual quality, while the VDM model is applied at low noise levels to improve likelihood (see \cref{fig:merged_model}). This configuration is motivated by prior findings indicating that likelihood is primarily influenced by denoising at low noise levels, whereas perceptual fidelity benefits from accurate denoising at high noise levels~\citep{kim2021soft,zheng2023improved}.

Let $\gamma^{\mathrm{EDM}}_t$ and $\gamma^{\mathrm{VDM}}_t$ denote the respective training noise schedules (negative log-SNR) of EDM and VDM. We introduce a switching time $\eta \in [0,1]$ and define a target process with noise schedule $\gamma_t$ such that:
\[
    \gamma_0 = \gamma^{\mathrm{VDM}}_0 < \gamma^{\mathrm{EDM}}_0 \leq \gamma_\eta \leq \gamma^{\mathrm{VDM}}_1 < \gamma^{\mathrm{EDM}}_1 = \gamma_1 \ .
\]
The EDM and VDM models were trained over $\gamma^{\mathrm{VDM}} \in[-13.3,5]$ and $\gamma^{\mathrm{EDM}} \in[-12.43,8.764]$, respectively, implying that the merged model operates over the combined range $\gamma \in [-13.3,8.764]$.
For simplicity, we adopt a linear schedule, following \citet{kingma2021variational}:
\begin{equation}
    \gamma_t \coloneqq \gamma^{\mathrm{VDM}}_0 + t \; (\gamma^{\mathrm{EDM}}_1 - \gamma^{\mathrm{VDM}}_0)\ ,\qquad t\in[0,1] \ .
    \label{eq:merged_gamma}
\end{equation}
This formulation constrains the feasible switching time $\eta$ to the interval
\begin{align}
    \eta_{\min} = \frac{\gamma^{\mathrm{EDM}}_0 - \gamma_0^{\mathrm{VDM}}}{\gamma^{\mathrm{EDM}}_1 - \gamma^{\mathrm{VDM}}_0} \approx 0.0394 \ , \qquad\eta_{\max} = \frac{\gamma^{\mathrm{VDM}}_1 - \gamma_0^{\mathrm{VDM}}}{\gamma^{\mathrm{EDM}}_1 - \gamma^{\mathrm{VDM}}_0} \approx 0.8294 \ .
    \label{eq: eta min max}
\end{align}
In our experiments, we vary the switching thresholds within the range $\eta \in [\eta_{\min}, \eta_{\max}]$, thereby restricting the denoising process to noise levels covered by both models.
All models use the variance-preserving (VP) formulation, i.e., $\alpha_t^2 + \sigma_t^2 = 1$, and operate directly in pixel space.

We report the performance of the original EDM and VDM models as baselines and further include results from previously published approaches to provide a comprehensive evaluation. The base models are evaluated under their native $\gamma$ ranges. Our experiments are conducted on CIFAR-10~\citep{krizhevsky2009learning} and ImageNet32~\citep{deng2009imagenet}. For consistency with prior work, we adopt the original ImageNet32 variant~\citep{van2016pixel} and denote it with an asterisk ($^*$) in our comparisons, as some other works use the updated official version~\citep{chrabaszcz2017downsampled}.
Further experimental details are provided in \cref{appendix_sec:impl_details}.

Sample quality is evaluated using the Fréchet Inception Distance (FID)~\citep{heusel2017gans} on 50k generated samples following the EDM evaluation protocol of~\citet{karras2022elucidating}. We also report test log-likelihood in bits per dimension (BPD) using two methods. The first estimates the standard variational lower bound \eqref{eq:def_vlb_likelihood}. The second computes the exact log-likelihood of $\z_0$ by integrating the PF~ODE~\eqref{eq:background:pf_ode} while tracking the log-density with \cref{eq:def_ode_likelihood}, and subsequently applies truncated normal dequantization~\citep{zheng2023improved}. This yields the bound~\eqref{eq: ODE ELBO} which is tighter than~\eqref{eq:def_vlb_likelihood}.

\subsection{Results}

We now present experimental results evaluating our merged diffusion framework, focusing on how the threshold parameter $\eta$ mediates the trade-off between likelihood and perceptual image quality.

\begin{table}
    \centering
    \caption{Test likelihood (ODE) in bits/dimension (BPD) on CIFAR-10 and ImageNet32.}
    \label{tab:results_ode_tn_horizontal}
    \resizebox{\linewidth}{!}{
    \begin{tabular}{l *{13}{c}}
        \toprule
        &  & \multicolumn{12}{c}{\textbf{Threshold}} \\
        \cmidrule{3-14}
         & & \textbf{EDM} & \textbf{$\bigeta_{min}$} & \textbf{0.1} & \textbf{0.2} & \textbf{0.3} & \textbf{0.4} & \textbf{0.5} & \textbf{0.6} & \textbf{0.7} & \textbf{0.8} & \textbf{$\bigeta_{max}$} & \textbf{VDM} \\
        \midrule
         \textbf{CIFAR-10} & \textbf{NLL} & 3.21 & 3.09 & 2.83 & 2.69 & 2.63 & \textbf{2.62} & 2.62 & 2.63 & 2.63 & 2.63 & 2.64 & 2.64 \\
        & \textbf{NFE} & 204 & 232 & 234 & 236 & 253 & 259 & 254 & 251 & 257 & 273 & 274 & 248 \\
         \midrule
         \textbf{ImageNet32} & \textbf{NLL} & 4.04 & 3.96 & 3.80 & 3.76 &  3.74 & 3.72 & 3.72 & \textbf{3.72} & 3.72 & 3.72 & 3.72 & 3.72 \\
        & \textbf{NFE} & 195 & 185 & 192 & 186 & 180 & 180 & 196 & 210 & 220 & 232 & 236 & 205 \\
        \bottomrule
    \end{tabular}
    }
\end{table}
\begin{table}
    \centering
    \caption{FID@50k on CIFAR-10 and ImageNet32 with deterministic (ODE) sampling.}
    \label{tab:results_fid}
    \resizebox{\linewidth}{!}{
    \begin{tabular}{l *{13}{c}}
        \toprule
        &  & \multicolumn{12}{c}{\textbf{Threshold}} \\
        \cmidrule{3-14}
         & & \textbf{EDM} & \textbf{$\bigeta_{min}$} & \textbf{0.1} & \textbf{0.2} & \textbf{0.3} & \textbf{0.4} & \textbf{0.5} & \textbf{0.6} & \textbf{0.7} & \textbf{0.8} & \textbf{$\bigeta_{max}$} & \textbf{VDM} \\
        \midrule
         \textbf{CIFAR-10} & \textbf{FID} & 2.02 & 2.04 & 2.05 & 2.03 & \textbf{2.01} & 2.14 & 2.82 & 4.75 & 6.86 & 7.67 & 7.73 & 9.37 \\
        & \textbf{NFE} & 125 & 145 & 147 & 159 & 169 & 173 & 193 & 221 & 239 & 226 & 238 & 206 \\
         \midrule
        \textbf{ImageNet32} & \textbf{FID} & 7.38 & 7.43 & 7.44 & 	7.39 & 7.26 & 6.98 & \textbf{6.58} & 6.72 & 7.15 & 7.15 & 7.11 & 9.85 \\
        & \textbf{NFE} & 120 & 140 & 144 & 150 & 166 & 169 & 180 & 204 & 207 & 189 & 189 & 158  \\
        \bottomrule
    \end{tabular}
    }
\end{table}

\paragraph{Quantitative evaluation.}
We investigate how the threshold parameter $\eta$ in our merged model governs the balance between data likelihood and perceptual image quality. \cref{tab:results_ode_tn_horizontal} reports negative log-likelihood (NLL) in bits per dimension (BPD) computed using the bound in \cref{eq: ODE ELBO}, which relies on the PF~ODE \cref{eq:background:pf_ode} and the truncated normal dequantization method proposed by \citet{zheng2023improved}. The corresponding FID scores for unconditional image generation using ODE-based sampling are reported in \Cref{tab:results_fid}.
We vary $\eta$ from $\eta_{\min}$ to $\eta_{\max}$ (see \cref{eq: eta min max}).

\begin{wrapfigure}[25]{r}{0.61\textwidth}
    \centering
    \vspace{-10pt}
    \includegraphics[width=\linewidth]{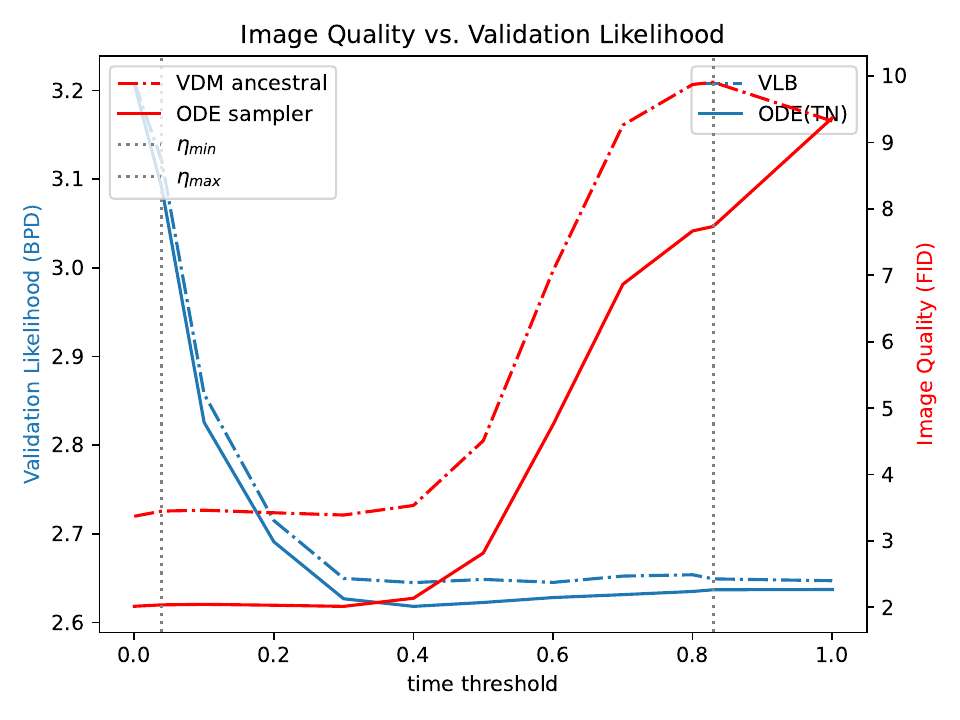}
    \caption{Likelihood--quality trade-off on CIFAR-10. Likelihood is measured in BPD using PF ODE integration with truncated normal dequantization (\emph{ODE (TN)}) or the variational lower bound (\emph{VLB}). Perceptual quality is measured with FID with both deterministic (\emph{ODE sampler}) and stochastic (\emph{VDM ancestral}) integration. The x-axis corresponds to the switching threshold $\eta$ between the models. The EDM and VDM base models correspond to $\eta = 0$ and $\eta = 1$, respectively.}
    \label{fig:evaluation_plot}
\end{wrapfigure}

\Cref{fig:evaluation_plot} visualizes the effect of $\eta$ on both likelihood and perceptual quality. For likelihood, we report both the ODE-based bound \eqref{eq: ODE ELBO} and the variational bound \eqref{eq:def_vlb_likelihood} which does not require ODE integration. For perceptual quality, we include FID scores obtained from deterministic sampling with an adaptive step size ODE solver, and from stochastic sampling with 256 sampling steps. For ease of interpretation, in \cref{fig:evaluation_plot}, we denote pure EDM as $\eta=0$ and pure VDM as $\eta=1$, although these values are not technically realizable in the merged model.

Varying $\eta$ produces a clear and consistent trade-off between likelihood and perceptual quality. On CIFAR-10, the best overall operating point occurs at $\eta = 0.3$. Increasing $\eta$ to $0.4$ further improves
the likelihood beyond the VDM baseline, with only a slight degradation in FID (2.02 to 2.14). On ImageNet32 (see \cref{fig:evaluation_all_imagenet32} in \cref{appendix_sec:additional_results}), $\eta = 0.5$ matches the VDM baseline in likelihood while surpassing EDM in FID.
\textbf{Overall, these results indicate that a single threshold $\eta$ can outperform both base models across metrics, demonstrating that our approach effectively breaks the apparent trade-off between likelihood and perceptual quality.}

To contextualize these results, \cref{tab:summary_result} compares our method with other approaches designed to achieve both high likelihood and strong perceptual quality. For likelihood evaluation, we use PF–ODE likelihood estimation with truncated normal dequantization, while image quality is assessed using ODE-based sampling.
Our approach outperforms Soft Truncation~\citep{kim2021soft}, which also seeks to balance these objectives. Consistency Trajectory Models~\citep[CTM,][]{kim2023consistency} improve both metrics by combining multiple loss functions, including GAN-based objectives, and data augmentation. In contrast, our method relies solely on standard denoising objectives. A broader comparison is presented in \cref{tab:full_result} (\cref{appendix_sec:additional_results}), which includes models achieving state-of-the-art results on either metric, providing a reference for current performance limits.
While we do not claim state-of-the-art performance, our experiments show that merging two pretrained diffusion models---one optimized for perceptual image quality and the other for likelihood---consistently improves both metrics compared to either model used independently.

\begin{table}
    \centering
    \caption{Comparison of our method with prior approaches targeting both high likelihood and strong perceptual quality. Unless otherwise noted, NLL is evaluated using truncated normal dequantization and PF ODE integration. Alternative settings are indicated as follows: Uniform Deq.$^\dagger$, Variational Deq.$^\ddagger$, VLB$^\vee$, Data Augmentation$^\uplus$., ImageNet32 (old version)$^\ast$.}
    \label{tab:summary_result}
    \resizebox{\linewidth}{!}{
    \begin{tabular}{l c c c c c c}
        \toprule
        \textbf{Model} & \multicolumn{3}{c}{\textbf{CIFAR-10}} & \multicolumn{3}{c}{\textbf{ImageNet32}} \\
         &  NLL$(\downarrow)$ & FID$(\downarrow)$ & NFE & NLL$(\downarrow)$ & FID$(\downarrow)$ & NFE \\
        \midrule
        \textbf{Base Models} \\
        VDM~\citep{kingma2021variational} & 2.65$^\vee$ & 7.41 & -  & 3.72$^{\ast\vee}$ & - & - \\
        EDM (w/ Heun Sampler)~\citep{karras2022elucidating} & - & 1.97 & 35 & - & - & - \\
        \midrule
        \textbf{Focused on both FID and NLL} \\
        Soft Truncation~\citep{kim2021soft} & 3.01$^\dagger$ & 3.96 & -  & 3.90$^{\ast\dagger}$ & 8.42$^\ast$ & - \\
        CTM ($\uplus$ - randomflip)~\citep{kim2023consistency} & 2.43$^\dagger$ & 1.87 & 2 & - & - & - \\
        \midrule
        \textbf{Ours} \\
        VDM (our evaluation, $\gamma\in $[-13.3, 5]) & 2.64/2.66$^\vee$ & 9.37 & 206 & 3.72$^\ast$/3.72$^{\ast\vee}$ & 9.85$^\ast$ & 158 \\
        EDM (our evaluation, $\gamma\in $[-12.43, 8.764]) & 3.21 & 2.02 & 125 & 4.04$^\ast$ & 7.38$^\ast$ & 120 \\
        Ours NLL ($\eta=0.4$, CIFAR-10) & {2.62} & 2.14 & 173 & - & - & - \\
        Ours ($\eta=0.3$, CIFAR-10) & {2.63} & {2.01} & 169 & - & - & - \\
        Ours ($\eta=0.5$, ImageNet32) & - & - & - & {3.72}$^\ast$ & {6.58}$^\ast$ & 180 \\
        \bottomrule
    \end{tabular}
    }
\end{table}

\begin{wrapfigure}[27]{r}{0.5\textwidth}
    \centering
    \vspace{-20pt}
    \includegraphics[width=\linewidth]{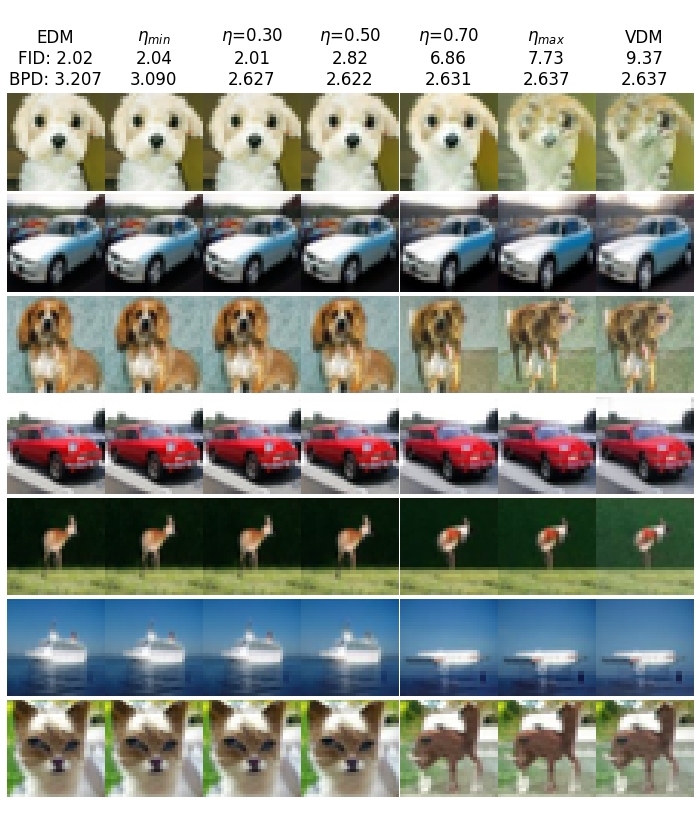}
    \vspace{-20pt}
    \caption{Qualitative comparison on CIFAR-10 using the ODE sampler. Each row starts from the same noise sample $\z_1$, while columns vary the threshold $\eta$ in the merged model. The model follows EDM dynamics up to time $\eta$ and then switches to VDM for $t<\eta$. Increasing $\eta$ triggers an earlier switch, improving likelihood but gradually reducing perceptual fidelity.}
    \label{fig:transition_plot}
\end{wrapfigure}

\paragraph{Qualitative evaluation.} 
\Cref{fig:transition_plot} presents qualitative results on CIFAR-10 obtained using the ODE sampler. In this setting, the score term $\nabla \log q(\z_t)$ in the PF~ODE \eqref{eq:background:pf_ode} is replaced by the learned score network $\stheta$, and the ODE is integrated backward in time to generate samples. For each row in the figure, we use the same Gaussian latent sample $\z_1 \sim \mathcal{N}(\0, \I)$ and vary only the switching threshold~$\eta$ indicated in the column headers. The merged model integrates the ODE backward and transitions from the pretrained EDM model to the pretrained VDM model at the specified threshold~$\eta$. Consequently, all trajectories are identical up to the switching point.

When sampling exclusively from the EDM model (leftmost column), we observe visually high-quality samples but relatively poor likelihood. As the switch occurs earlier in the denoising trajectory (i.e., as~$\eta$ increases), the likelihood improves while perceptual quality remains largely unchanged, up to a point where excessive reliance on the VDM component begins to degrade image fidelity. This trend aligns precisely with the intended behavior of our approach: early EDM stages produce realistic, high-quality structures, while the subsequent VDM refinement enhances likelihood with only minor perceptual alterations. Notably, despite differences in architecture, training objectives, and weighting of the ELBO, the two base models frequently yield remarkably similar generated samples from the same initial noise, both individually and when combined within the merged framework.

\section{Related Work}
\label{sec:related}

\textbf{Likelihood experts.} \ 
Several methods focus on improving likelihood. VDM~\citep{kingma2021variational} and ScoreFlow~\citep{song2021maximum} directly optimize (a bound on) the data log-likelihood. i-DODE~\citep{zheng2023improved} introduces \textit{velocity prediction} and proposes an improved likelihood estimation technique. Other works~\citep{sahoo2023diffusion, nielsen2023diffenc, bartosh2024neural} explore learnable forward processes, whereas our study focuses on standard diffusion models with fixed linear forward noise schedules.

\textbf{Sample quality experts.} \ 
Many studies improve the perceptual quality of generated samples by introducing better or more efficient samplers~\citep{song2020denoising, song2020score, lu2022dpm, zheng2023dpm, zhao2024unipc, karras2022elucidating, zhou2024score}, addressing exposure bias~\citep{ning2023elucidating}, or applying alternative loss weighting strategies~\citep{kingma2024understanding, ho2020denoising}. GMEM~\citep{tang2024generative} enhances both quality and efficiency by incorporating an external memory bank into a transformer-based model, achieving state-of-the-art FID on CIFAR-10. PaGoDA~\citep{kim2024pagoda}, a distillation-based approach, achieves the best known FID on ImageNet32. In this work, we focus on UNet-based diffusion models trained with simpler objectives such as \textit{noise prediction}, and exclude distillation-based methods from our scope.

\textbf{Experts on both metrics.} \ 
Soft Truncation~\citep{kim2021soft} proposes a training strategy that softens fixed truncation into a random variable, adjusting loss weighting across diffusion times to address the likelihood–quality trade-off.
While aligned in motivation with our work, their approach requires training from scratch. In contrast, our method directly leverages existing pretrained models. CTM~\citep{kim2023consistency} uses a combination of loss terms, including an additional GAN loss, along with data augmentation to improve both metrics. In contrast, we address the trade-off from a different perspective, by merging experts trained with the standard denoising objective.

\textbf{Mixture-of-Experts.} \ 
Mixture-of-Experts (MoE) frameworks have been applied to diffusion models in contexts such as zero-shot text-to-image generation~\citep{balaji2022ediff, feng2023ernie} and controllable image synthesis~\citep{bar2023multidiffusion}. More recently, MDM~\citep{kang2024local} proposed a MoE strategy where each expert is trained on a specific time interval. While effective, their method employs identical architectures across experts and primarily targets training efficiency and sample quality. To the best of our knowledge, we are the first to address this trade-off by merging pretrained experts specialized separately in likelihood and sample quality.

\section{Conclusion}
\label{sec:conclusion}

We proposed a simple yet effective approach for merging pretrained diffusion or flow models to mitigate the trade-off between likelihood and perceptual image quality. By switching between an expert optimized for perceptual fidelity and another optimized for data likelihood, the hybrid model achieves consistent improvements across both objectives. On CIFAR-10 and ImageNet32, it matches or surpasses the performance of its individual components, demonstrating that complementary models can be combined to yield stronger generative behavior without retraining.

A key advantage of the method is its complete reliance on existing pretrained models, requiring no fine-tuning or additional supervision. Its simplicity, however, comes with limitations: performance depends on the characteristics of the merged models, and the optimal switching threshold must be determined empirically. Furthermore, our experiments are restricted to pixel-space diffusion models, leaving room for adaptation to other architectures and training regimes.

Future research may explore automated or learned switching mechanisms, integration with advanced samplers, and extensions to latent or consistency-based diffusion models. Overall, this work highlights model merging as a lightweight yet powerful tool for enhancing pretrained diffusion systems, suggesting new directions for efficient and modular generative modeling.

\bibliography{iclr2025_delta}
\bibliographystyle{iclr2025_delta}

\clearpage

\appendix

\section{Experimental Details}
\label{appendix_sec:impl_details}

\paragraph{Datasets.}
We conduct experiments on the CIFAR-10 and ImageNet32 datasets. For CIFAR-10~\citep{krizhevsky2009learning}, we use the standard version distributed through PyTorch.
For ImageNet32, we use its original version~\citep{deng2009imagenet,van2016pixel}, which is no longer officially distributed but has been made available by \citet{zheng2023improved}. Since some prior works report results using the newer official release of ImageNet32~\citep{chrabaszcz2017downsampled}, we indicate, for each baseline, which dataset version their results correspond to.

\paragraph{Models and training.}
We use the publicly available EDM checkpoint for CIFAR-10~\citep{krizhevsky2009learning}.\footnote{\url{https://nvlabs-fi-cdn.nvidia.com/edm/pretrained/edm-cifar10-32x32-uncond-vp.pkl}} For VDM, we train the PyTorch re-implementation\footnote{\url{https://github.com/addtt/variational-diffusion-models}} based on the architecture described in~\citet{kingma2021variational}. The model is trained for 10 million steps on 8×A100 (40GB) GPUs, with no data augmentation, a fixed linear $\gamma$ schedule, and a batch size of 128. The resulting model achieves 2.64 BPD on the test set (PF~ODE likelihood with TN dequantization) and 2.66 BPD under VLB evaluation.

For ImageNet32, VDM is trained similarly to CIFAR-10, but with 256 channels and a total batch size of 512 
for 2 million steps, following~\citet{kingma2021variational}.
Since no pretrained EDM model is publicly available for ImageNet32, we train one using the official EDM code, with parameters \texttt{--cond 0 --arch ddpmpp --duration 1000}. Training is performed for 1000M images
with a total batch size of 1024. No hyperparameter tuning was performed in this case.

\paragraph{Evaluation.}
To compute the divergence of the PF~ODE vector field when evaluating $\log \ptheta (\z_0)$ in \cref{eq: ODE ELBO}, we use the Skilling--Hutchinson trace estimator~\citep{skilling1989eigenvalues, hutchinson1989stochastic}:
\begin{equation}
\nabla \cdot \htheta (\z_t, t) = \operatorname{tr}\left(\frac{\partial\, \htheta\!\left(\z_t, t\right)}{\partial \z_t}\right)=\E_{p(\bepsilon)}\left[\bepsilon^{\top} \; \frac{\partial\, \htheta\!\left(\z_t, t\right)}{\partial \z_t} \; \bepsilon\right]
\end{equation}
where $\frac{\partial\, \htheta\!\left(\z_t, t\right)}{\partial \z_t}$ is the Jacobian of $\htheta$, and $\bepsilon$ is a random variable such that $\E_{p(\bepsilon)}[\bepsilon]=\0$ and $\text{Cov}_{p(\bepsilon)}[\bepsilon]=\I$. We use the Rademacher distribution for $p(\bepsilon)$ and use the RK45 ODE solver~\citep{DORMAND198019} with \texttt{atol=1e-5} and \texttt{rtol=1e-5}, following prior work~\citep{sahoo2023diffusion, song2020score, zheng2023improved}.
For all evaluations, we use Exponential Moving Average (EMA) weights. FID scores are computed following~\citet{karras2022elucidating}, with reference statistics calculated from the training sets.

\section{Additional results and visualizations}
\label{appendix_sec:additional_results}

\begin{figure}
    \centering
    \begin{minipage}[t]{1\textwidth}
        \centering
        \includegraphics[width=\linewidth]{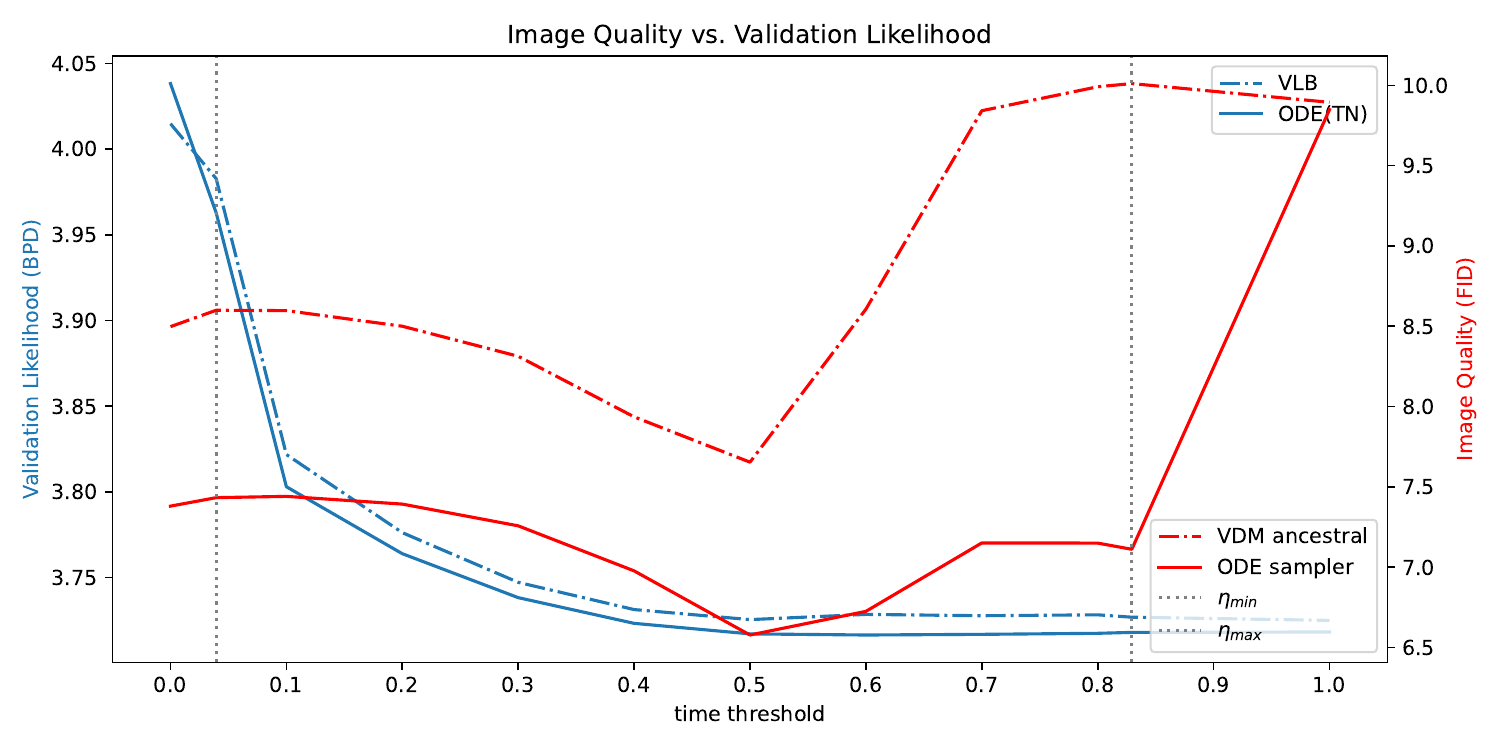}
        \caption{Likelihood--quality trade-off on ImageNet32. Likelihood is measured in BPD using PF ODE integration with truncated normal dequantization (\emph{ODE (TN)}) or the variational lower bound (\emph{VLB}). Perceptual quality is measured with FID with both deterministic (\emph{ODE sampler}) and stochastic (\emph{VDM ancestral}) integration. The x-axis corresponds to the switching threshold $\eta$ between the models. The EDM and VDM base models correspond to $\eta = 0$ and $\eta = 1$, respectively.}
        \label{fig:evaluation_all_imagenet32}
    \end{minipage}
\end{figure}

\cref{tab:results_vlb} reports likelihood evaluations of the merged model using the Variational Lower Bound (VLB) from \cref{eq:def_vlb_likelihood}. We report the mean and standard deviation over 10 runs.
\cref{fig:evaluation_all_imagenet32} shows the trade-off between likelihood and perceptual quality on ImageNet32. For likelihood we use both the ODE (\cref{eq: ODE ELBO}, with truncated normal dequantization) and SDE (\cref{tab:results_vlb}) bounds, and for evaluating perceptual quality we use both the ODE sampler and ancestral sampling. The EDM and VDM baselines correspond to $\eta = 0.0$ and $\eta = 1.0$, respectively, and do not involve expert switching. Note again here it's improper to define them as $\eta = 0.0$ and $\eta = 1.0$ but convenient for plotting.

\begin{table}
    \centering
    \captionsetup{justification=centering}
        \centering
        \renewcommand{\arraystretch}{1.2}
        \setlength{\tabcolsep}{11pt}

        \caption{VLB evaluation in terms of bits per dimension (BPD) on CIFAR-10 and ImageNet32.}
        \label{tab:results_vlb}
        \begin{tabular}{l c c c c}
            \toprule
            \textbf{Threshold} & \multicolumn{2}{c}{\textbf{CIFAR-10}} & \multicolumn{2}{c}{\textbf{ImageNet32}} \\
             &  Mean$(\downarrow)$ & Std & Mean$(\downarrow)$ & Std \\
            \midrule
            0.0    & 3.21 & 0.06  & 4.01 & 0.004 \\
            $\eta_{\min}$   & 3.12 & 0.005 & 3.98 & 0.004 \\
            0.1    & 2.86 & 0.009 & 3.82 & 0.004 \\
            0.2    & 2.71 & 0.008 & 3.78 & 0.004 \\
            0.3    & 2.65 & 0.007 & 3.75 & 0.005 \\
            0.4    & 2.64 & 0.008 & 3.73 & 0.005 \\
            0.5    & 2.65 & 0.009 & 3.73 & 0.004 \\
            0.6    & 2.65 & 0.007 & 3.73 & 0.005 \\
            0.7    & 2.65 & 0.008 & 3.73 & 0.007 \\
            0.8    & 2.65 & 0.009 & 3.73 & 0.005 \\
            $\eta_{\max}$   & 2.65 & 0.008 & 3.73 & 0.002 \\
            1.0    & 2.65 & 0.005 & 3.73 & 0.004 \\
            \bottomrule
        \end{tabular}
\end{table}

\begin{figure}
    \centering
    \includegraphics[scale=0.45]{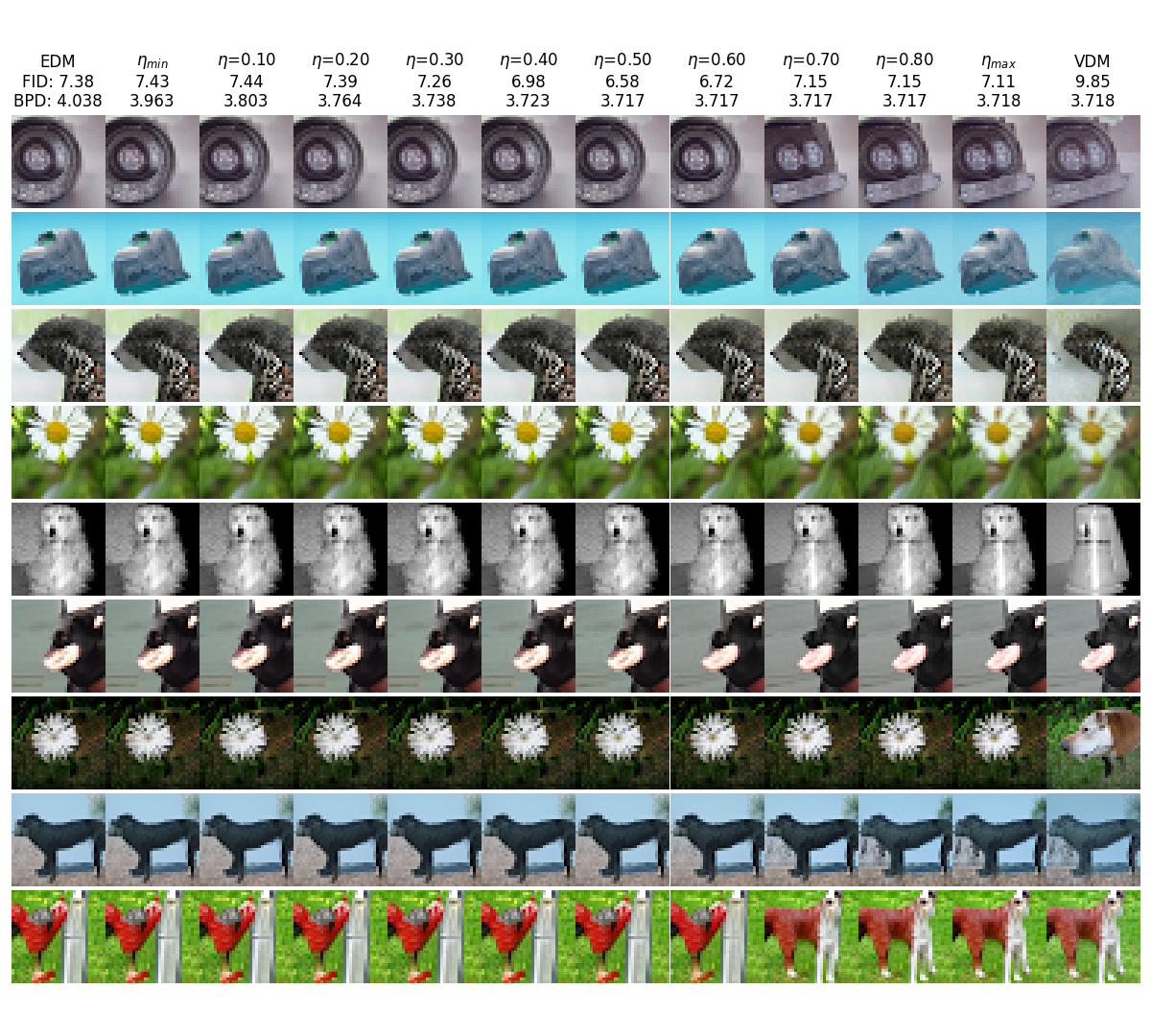}
    \caption{Generated images from our merged model using different thresholds $\eta$ on ImageNet32 dataset.}
    \label{fig:generated_samples_diff_thresholds_in32}
\end{figure}

\cref{fig:generated_samples_diff_thresholds_in32} shows generated samples from our merged model on ImageNet32. For each row, we fix a noise sample $\z_1 \sim p(\z_1)$ and generate samples with the merged model varying the switching threshold~$\eta$.
In the leftmost column of the figure (corresponding to EDM), the samples exhibit excellent perceptual quality but poor likelihood. As $\eta$ increases, likelihood improves while image quality remains largely unchanged. Around $\eta = 0.5$, the model achieves the same likelihood as the VDM expert while surpassing the EDM baseline in FID.

\begin{figure}
    \centering
    \includegraphics[width=0.75\linewidth]{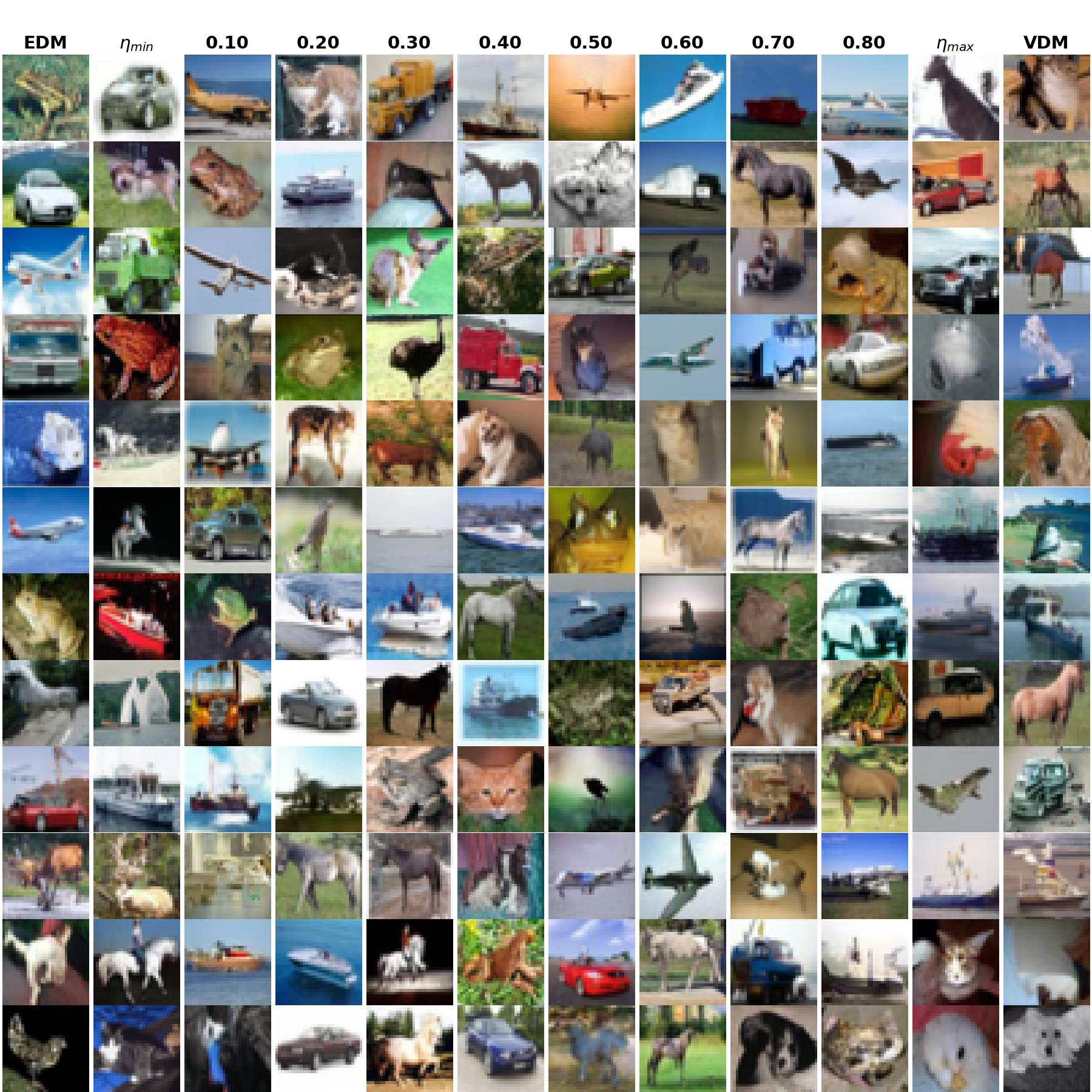}
    \caption{Random samples on CIFAR-10 using ODE sampler (with different $\eta$).}
    \label{fig:cifar10-ode}
\end{figure}

\begin{figure}
    \centering
    \includegraphics[width=0.75\linewidth]{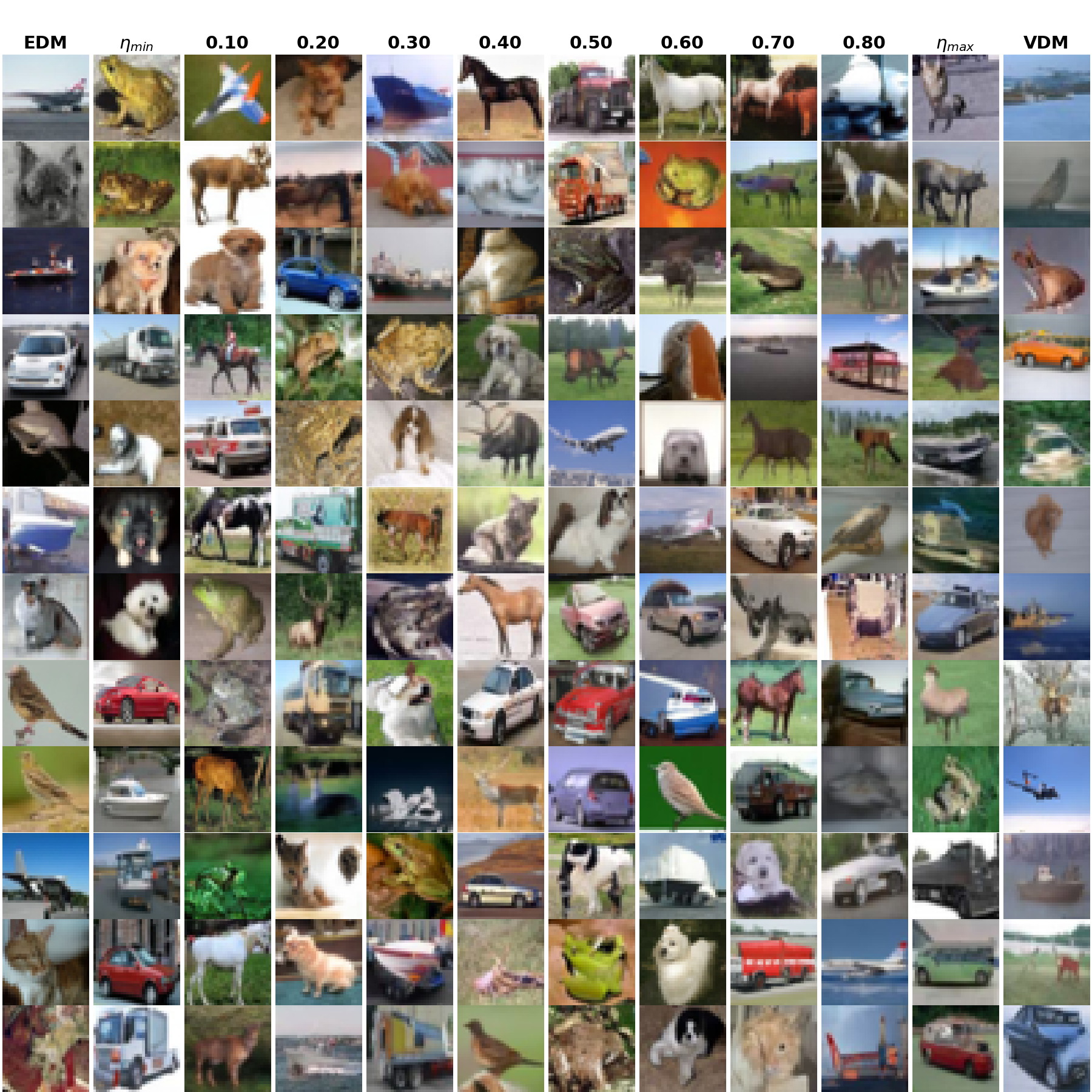}
    \caption{Random samples on CIFAR-10 using VDM ancestral sampler (with different $\eta$).}
    \label{fig:cifar10-vdm}
\end{figure}

\begin{figure}
    \centering
    \includegraphics[width=0.75\linewidth]{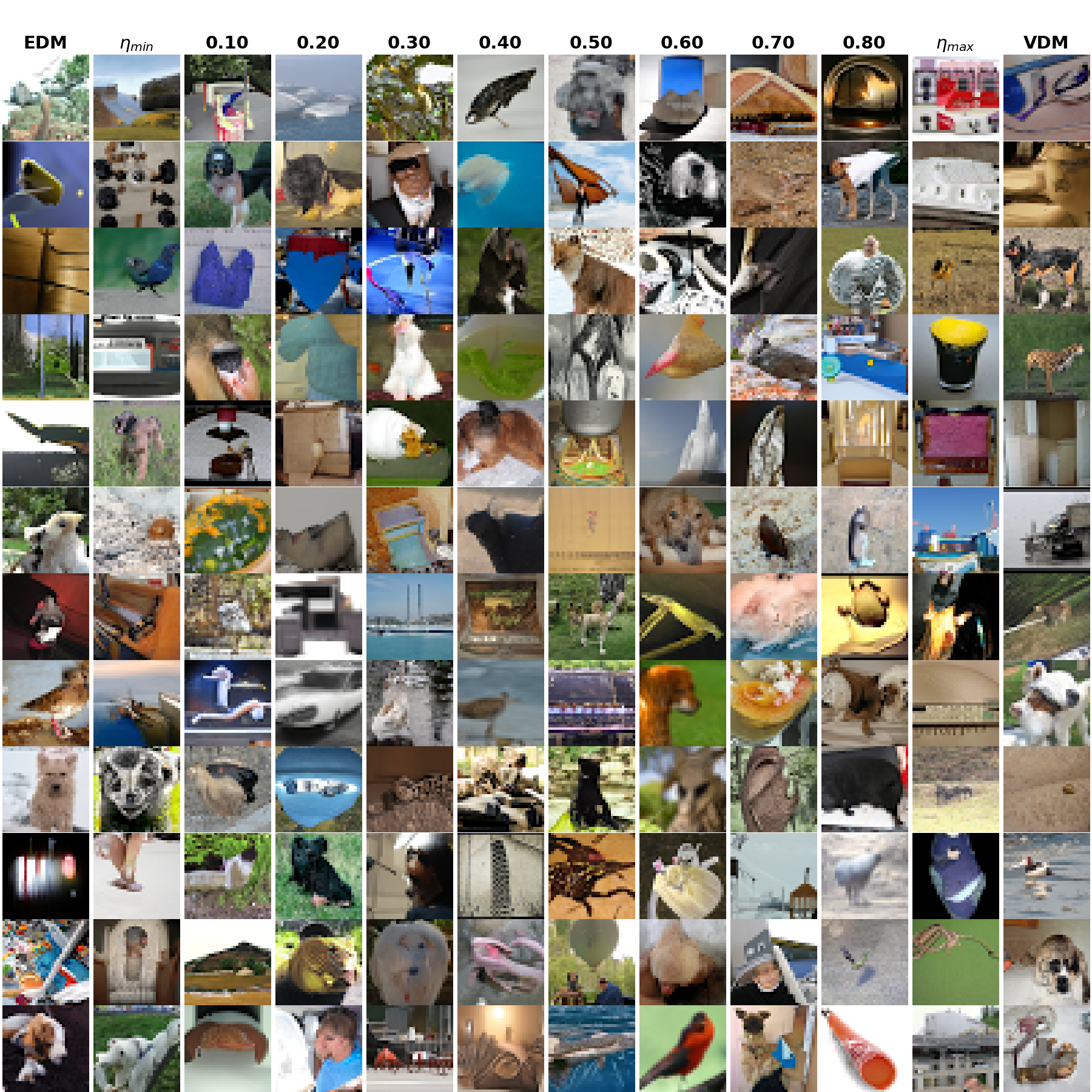}
    \caption{Random samples on ImageNet32 using ODE sampler  (with different $\eta$).}
    \label{fig:imagenet32-ode}
\end{figure}

\begin{figure}
    \centering
    \includegraphics[width=0.75\linewidth]{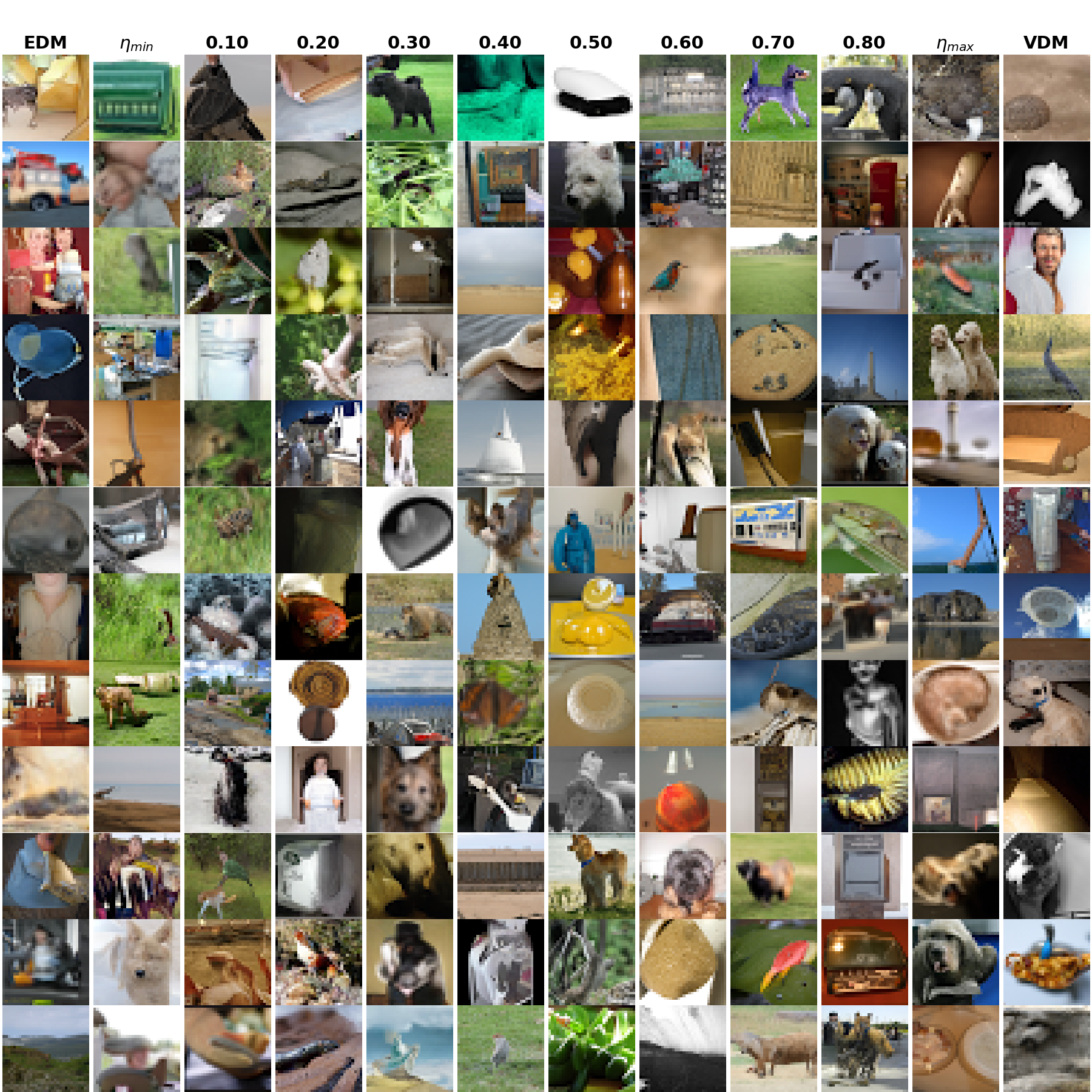}
    \caption{Random samples on ImageNet32 using VDM ancestral sampler  (with different $\eta$).}
    \label{fig:imagenet32-vdm}
\end{figure}

In \cref{fig:cifar10-ode,fig:cifar10-vdm,fig:imagenet32-ode,fig:imagenet32-vdm}, we present randomly generated samples (without fixed noise samples $\z_1$) from the baseline models and the merged models (across a range of $\eta$ values), both for CIFAR-10 and ImageNet32, using both deterministic and stochastic sampling.

\begin{table}
    \centering
    \caption{Extended version of \cref{tab:summary_result} with additional methods from the literature. Unless otherwise noted, NLL is evaluated using truncated normal dequantization and PF ODE integration. Alternative settings are indicated as follows: Uniform Deq.$^\dagger$, Variational Deq.$^\ddagger$, VLB$^\vee$, Data Augmentation$^\uplus$, ImageNet32(old version)$^\ast$.}
    \label{tab:full_result}
    \resizebox{\linewidth}{!}{
    \begin{tabular}{l c c c c c c}
        \toprule
        \textbf{Model} & \multicolumn{3}{c}{\textbf{CIFAR-10}} & \multicolumn{3}{c}{\textbf{ImageNet32}} \\
         &  NLL$(\downarrow)$ & FID$(\downarrow)$ & NFE & NLL$(\downarrow)$ & FID$(\downarrow)$ & NFE \\
        \midrule
        \textbf{Base Models} \\
        VDM~\citep{kingma2021variational} & 2.65$^\vee$ & 7.41 & -  & 3.72$^{\ast\vee}$ & - & - \\
        EDM (w/ Heun Sampler)~\citep{karras2022elucidating} & - & 1.97 & 35 & - & - & - \\
        \midrule
        \textbf{Focused on both FID-NLL} \\
        Soft Truncation~\citep{kim2021soft} & 3.01$^\dagger$ & 3.96 & -  & 3.90$^{\ast\dagger}$ & 8.42$^\ast$ & - \\
        CTM ($\uplus$ - randomflip)~\citep{kim2023consistency} & 2.43$^\dagger$ & 1.87 & 2 & - & - & - \\
        ScoreSDE ($\uplus$ - randomflip)~\citep{song2020score} & 2.99$^\dagger$ & 2.92 & -  & - & - & - \\
        LSGM (FID)~\citep{vahdat2021score}& 3.43 & 2.10 & - & - & -  & - \\
        DDPM++ cont. (deep, sub-VP)~\citep{song2020score} & 2.99$^\dagger$ & 2.92 & - & - & - & - \\
        Reflected Diffusion Models~\citep{lou2023reflected} & 2.68 & 2.72 & - & 3.74 & - & - \\
        \midrule
        \textbf{Focused on FID} \\
        GMEM (Transformer-based)~\citep{tang2024generative} & - & 1.22 & 50 & - & - & - \\
        PaGoDA (distillation-based)~\citep{kim2024pagoda} & - & - & - & - & 0.79 & 1 \\
        SiD (distillation-based)~\citep{zhou2024score} & - & 1.923 & 1 & - & - & - \\
        ScoreFlow (VP, FID)~\citep{song2021maximum} & 3.04$^\ddagger$ & 3.98 & -  & 3.84$^{\ast\ddagger}$ & 8.34$^\ast$ & -  \\
        PNDM~\citep{liu2022pseudo} & - & 3.26 & - & - & -  & - \\
        \midrule
        \textbf{Focused on NLL} \\
        i-DODE (VP)~\citep{zheng2023improved} & 2.57 & 10.74 & 126 & 3.43/3.70$^\ast$ & 9.09 & 152 \\
        i-DODE (VP, $\uplus$)~\citep{zheng2023improved} & 2.42 & 3.76 & 215 & - & - & - \\
        Flow Matching~\citep{lipman2022flow} & 2.99$^\dagger$ & 6.35 & 142 & 3.53$^\dagger$ & 5.02 & 122 \\
        DiffEnc~\citep{nielsen2023diffenc} & 2.62$^\vee$ & 11.1 & - & 3.46$^\vee$ & - & - \\
        NDM ($\uplus$ - horizontalflip)~\citep{bartosh2023neural} & 2.70$^\dagger$ & - & - & 3.55 & - & - \\
        NFDM (Gaussian q, $\uplus$ - horizontalflip)~\citep{bartosh2024neural} & 2.49$^\dagger$ & 21.88 & 12 & 3.36 & 24.74 & 12 \\
        NFDM (non-Gaussian q, $\uplus$ - horizontalflip)~\citep{bartosh2024neural} & 2.48$^\dagger$ & - & - & 3.34 & - & - \\
        NFDM-OT($\uplus$ - horizontalflip)~\citep{bartosh2024neural} & 2.62$^\dagger$ & 5.20 & 12 & 3.45 & 4.11 & 12 \\
        ScoreFlow (deep, sub-VP, NLL)~\citep{song2021maximum} & 2.81$^\ddagger$ & 5.40 & -  & 3.76$^{\ast\ddagger}$ & 10.18$^\ast$ & -  \\
        Stochastic Interp.~\citep{albergo2022building} & 2.99$^\dagger$ & 10.27 & - & 3.48$^\dagger$ & 8.49 & - \\
        MuLAN (w/o IS $k=1$)~\citep{sahoo2023diffusion} & 2.59 & - & - & 3.71 & - & - \\
        MuLAN (w/ IS $k=20$)~\citep{sahoo2023diffusion} & 2.55 & - & - & 3.67 & - & - \\
        Improved DDPM ($L_{\text{vlb}}$)~\citep{nichol2021improved} & 2.94$^\vee$ & 11.47 & - & - & -  & - \\        FFJORD~\citep{grathwohl2018ffjord} & 3.4 & - & -  & - & - & - \\
        Improved DDPM ($L_{\text{vlb}}$)~\citep{nichol2021improved} & 2.94$^\vee$ & 11.47 & - & - & -  & - \\
        ARDM-Upscale 4 (autoregressive)~\citep{hoogeboom2021autoregressive} & 2.64 & - & - & - & - & -\\
        Efficient-VDVAE~\citep{hazami2022efficientvdvae} & 2.87$^\vee$ & - & - & 3.58 & -  & - \\
        DenseFlow-74-10~\citep{grcic2021densely} & 2.98$^\ddagger$ & 34.90 & -  & 3.63 & - & -  \\
        \midrule
        \textbf{Ours} \\
        VDM (our evaluation, $\gamma\in $[-13.3, 5])~\citep{kingma2021variational} & 2.64/2.66$^\vee$ & 9.37 & 206 & 3.72$^\ast$/3.72$^{\ast\vee}$ & 9.85$^\ast$ & 158 \\
        EDM (our evaluation, $\gamma\in $[-12.43, 8.764])~\citep{karras2022elucidating} & 3.21 & 2.02 & 125 & 4.04$^\ast$ & 7.38$^\ast$ & 120 \\
        Ours NLL ($\eta=0.4$, CIFAR-10) & {2.62} & 2.14 & 173 & - & - & - \\
        Ours ($\eta=0.3$, CIFAR-10) & {2.63} & {2.01} & 169 & - & - & - \\
        Ours ($\eta=0.5$, ImageNet32) & - & - & - & {3.72}$^\ast$ & {6.58}$^\ast$ & 180 \\
        \bottomrule
    \end{tabular}
    }
\end{table}

\end{document}